\documentclass[lnbip]{svmultln}
\usepackage{makeidx}  

\usepackage{graphicx}
\usepackage{xcolor}

\newcommand{\sectionname}{Sec.}

\begin{document}

\mainmatter              

\title{Agentic Business Process Management Systems}
\titlerunning{Agentic Business Process Management Systems}  
%
\author{Marlon Dumas\inst{1} \and Fredrik Milani\inst{1} \and David Chapela-Campa\inst{1}}
\authorrunning{Marlon Dumas et al.}   
%
\tocauthor{Marlon Dumas, Fredrik Milani, David Chapela-Campa}
\institute{
    University of Tartu, Tartu, Estonia\\
    \email{{marlon.dumas, fredrik.milani, david.chapela}@ut.ee}
}

\maketitle              


\begin{abstract}        
Since the early 90s, the evolution of the Business Process Management (BPM) discipline has been punctuated by successive waves of automation technologies. 
Some of these  technologies enable the automation of individual tasks, while others focus on orchestrating the execution of end-to-end processes.
The rise of Generative and Agentic Artificial Intelligence (AI) is opening the way for another such wave. However, this wave is poised to be different because it shifts the focus from automation to autonomy and from design-driven management of business processes to data-driven management, leveraging process mining techniques. This position paper, based on a keynote talk at the 2025 Workshop on AI for BPM, outlines how process mining has laid the foundations on top of which agents can sense process states, reason about improvement opportunities, and act to maintain and optimize performance. The paper proposes an architectural vision for Agentic Business Process Management Systems (A-BPMS): a new class of platforms that integrate autonomy, reasoning, and learning into process management and execution. The paper contends that such systems must support a continuum of processes, spanning from human-driven to fully autonomous, thus redefining the boundaries of process automation and governance.
\end{abstract}

\section{Introduction}

Over the past five decades, process-aware information systems~\cite{Dumas2005} have evolved from paper-driven systems supporting manual work and coordination into today's highly digitized and automated systems. Initially, workers relied on personal knowledge, paper documents, and individual judgment to complete tasks. Early business process automation waves introduced tools like standard operating procedures, checklists, and forms, followed by spreadsheets and collaborative software that organized work while still requiring human execution. 

A leap occurred with the advent of case management and workflow management systems that could track work items and maintain digital records~\cite{davenport1993process}. Soon after, Business Process Management Systems (BPMS) emerged, enabling organizations to design process models separately from their execution using notations like BPMN, while workflow engines automated routing between systems and people~\cite{dumas2018fundamentals}. Parallel developments included business rules management systems that separated decision-making logic from applications, enabling automatic processing of routine decisions, and robotic process automation (RPA) that mimicked human interactions with software applications via their user interfaces~\cite{van2018robotic}. Today’s generation of process execution systems integrate real-time event processing, APIs, and machine learning to achieve higher levels of automation, combining execution with continuous monitoring and optimization~\cite{janiesch2025process}.

The rise of Generative AI and Agentic AI are laying the ground for the next evolution in this field, by enabling the development of tools capable not only of executing predefined workflows, but also to dynamically generate new process variations, make contextualized decisions, and autonomously adapt workflows based on changing conditions and objectives~\cite{dumas2023ai}. Agentic AI differs from rule-based or script-based automation. Rule-based or script-based automation technology, such as RPA, are characterized by their deterministic nature. They follow a set of fixed, pre-defined rules. This capability makes them suitable for automating repetitive tasks, for which all possible scenarios can be anticipated and scripted. Agentic AI, on the other hand, can operate beyond a set of rules by sensing and reasoning about the current state of processes, and triggering next actions that align with performance objectives, within a set of constraints (also known as a frame~\cite{dumas2023ai}).

Agents are software entities that have the capability to sense (understand input), decide (analyse and determine what to do), and act (e.g.\ execute a task). Hence, agentic AI can emulate a human resource~\cite{acharya2025agentic}.
In the context of business process management and execution, agents can make use of process monitoring and process mining tools to sense the state of the process in real-time. Based on patterns extracted from historical executions, they can detect situations where certain actions need to be triggered, even if these actions deviate from predesigned execution flows. By leveraging rule-based or script-based automation tools, as well as automated planning techniques, AI agents can trigger actions to drive the execution of one or more processes in an adaptive manner.

Inspired by an earlier definition of AI-Augmented Business Process Management Systems\cite{dumas2023ai}, we define an Agentic Business Process Management System as a class of process-aware information systems that leverages agentic AI technology to enact business processes in such a way that: (1) the execution flows of a process are not (fully) pre-determined via predesigned rules, models or scripts; (2) adaptations to automated components of the process may not require explicit changes to the supporting software applications; and (3) improvement opportunities may be autonomously discovered, validated, and applied.

The rest of the paper outlines an architecture for an envisioned class of process management and execution systems, namely Agentic BPM system (A-BPMS). The paper argues that such systems need to support a range of processes, with different levels of autonomy embedded in them, all the way from largely manual or largely automated processes to almost-fully autonomous ones.


\section{From Process Mining to Agentic BPM Systems}
\label{sec:from-process-mining-to-augmented}

\begin{figure}[t]
    \centering
    \includegraphics[width=0.6\columnwidth]{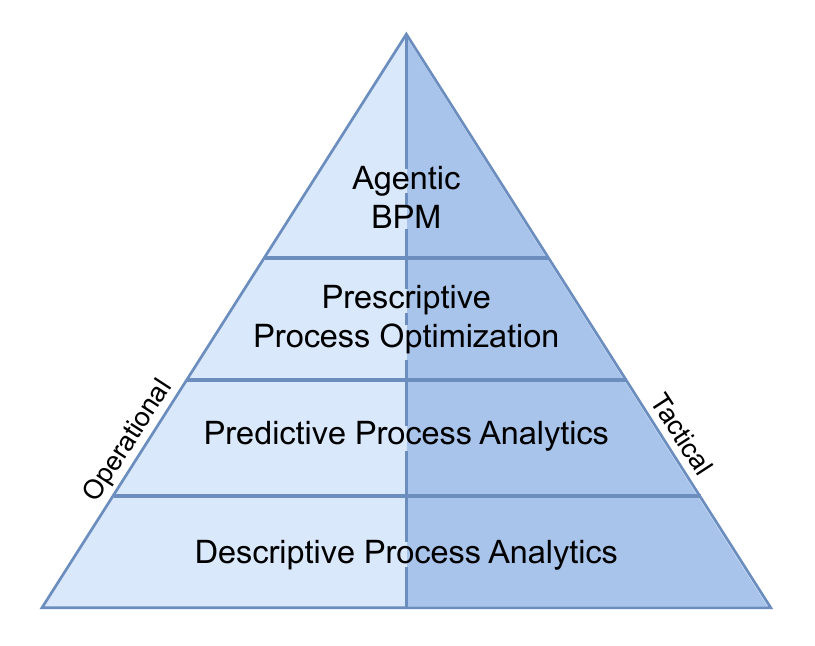}
    \vspace{-5pt}
    \caption{The Agentic BPM Pyramid: a classification of data-driven BPM approaches (adapted from~\cite{DBLP:journals/sosym/ChapelaCampaD23}).}
    \label{fig:bpm-pyramid}
\end{figure}

Underpinning an Agentic Business Process Management System (A-BPMS) is a collection of tools that allow for the transformation of data into insights and decisions.
This collection of data-driven techniques can be conceptualized in the form of a pyramid of capabilities, depicted in \figurename~\ref{fig:bpm-pyramid}.
In this pyramid, each layer builds on top of the layers below and serves as the basis for the ones above, and comprises techniques pertaining to two use cases:
\textit{i)} \emph{tactical}, aiming to inform managers in their business process change decisions, typically with timeframes of a few weeks to a few months between decision and change implementation;
and \textit{ii)} \emph{operational}, aiming to issue recommendations or trigger actions in the context of running cases, to improve their performance on a day-to-day basis.


The first layer, \emph{Descriptive Process Analytics}, comprises techniques focused on describing the current state of the process, generally addressing tactical use cases.
Within this layer, we find four main capabilities:
(1) \emph{Automated process discovery}, aiming to discover process models from data, in order to exhibit the behavior happening in the process, identify unexpected exceptions, and highlight potential wastes.
(2) \emph{Conformance checking}, aiming to analyze the behavior recorded in the event log against the behavior modeled by the process model, enabling the identification of deviations in the process and the comparison between the designed and the real process.
(3) \emph{Performance mining}, aiming to analyze one or more processes in terms of performance measures, such as cycle time or cost per case.
(4) \emph{Variant analysis}, aiming to identify positive and negative deviance in a process by comparing how the process is performed for different subsets of cases -- e.g., in different regions.

The second layer, \emph{Predictive process analytics}, moves beyond describing the current state to estimating the future state of the process.
The capabilities within this layer can be broadly classified into two categories:
(1) \emph{What-if digital process twins}, aiming to predict the impact of a (potential) process change at a macro-level.
By applying process mining, statistical analysis, and machine learning techniques to historical event logs, we can construct a digital process twin (DPT) capable of replicating the behavior of the process.
The DPT can then be used to estimate the impact that a change in the process might have on its performance metrics.
(2) \emph{Predictive process monitoring}, aiming to predict future states of a process.
Typically implemented through machine learning or deep learning techniques, these approaches can operate both 
\textit{i)} at the \emph{case level}, making predictions about individual cases in a process;
and \textit{ii)} at the \emph{process level}, predicting one or more process performance metrics across the entire process.


The third layer, \emph{Prescriptive Process Optimization}, focuses on prescribing actions that, based on the current state and future predictions, might increase (or decrease) the probability of certain (un)desired events occurring -- e.g., a loan offer being accepted by the customer.
These capabilities can be sub-categorized as follows:
(1) \emph{Automated process optimization}, aiming to recommend changes to a process in order to strike a tradeoff between competing performance indicators -- e.g., lowering cycle times while maintaining costs.
(2) \emph{Prescriptive process monitoring}, aiming to recommend actions in (near-)real-time to optimize the performance of a process. 
Existing techniques in this category are mainly designed to recommend actions to improve the performance at the \emph{case level}. 


Agentic BPM rests on top of the pyramid, illustrating the dependency of A-BPM systems on process intelligence. 
The aim of A-BPM systems is to autonomously manage and optimize processes to achieve the desired business outcomes, within the constraints and boundaries set by managers.
This objective cannot be achieved without the insights and recommendations of the data-driven techniques from the layers below.
Within this layer, we find two main categories of systems:
(1) \emph{Automated Systems} achieve the independence from human interaction through techniques that define, a priori, the entire set of decisions that can be made within the process -- e.g., rule-based orchestration and execution.
Although an automated system can theoretically function autonomously, the possibility of unforeseen situations with no predefined procedures to follow may force the system to either halt execution or depend on additional human input.
(2) \emph{Autonomous Systems} are composed of (AI) agents with the power to orchestrate and manage the process without human supervision.
In an autonomous system, humans may act as supervisors, (proactively) intervening only in order to prevent undesired consequences.

\section{Architecture of Agentic BPM Systems}

\begin{figure}[t]
    \centering
    \includegraphics[width=\columnwidth]{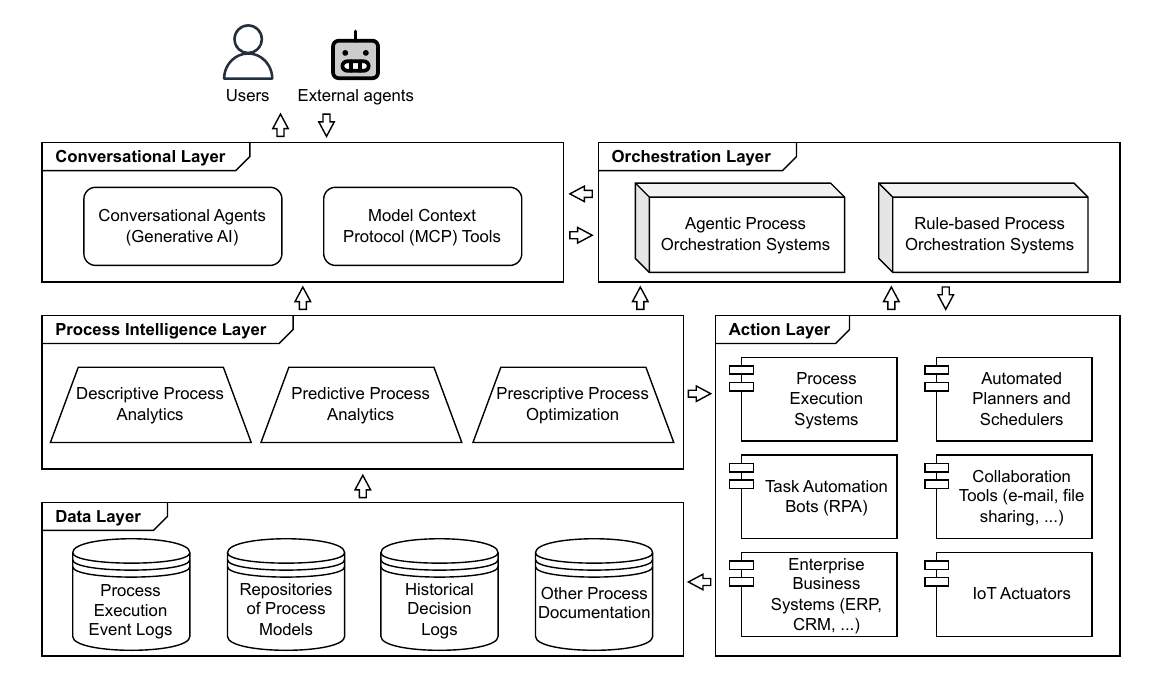} 
    \vspace{-15pt}
    \caption{Layered architecture of an Agentic BPM System.}
    \label{fig:abpms-architecture}
\end{figure}

We propose an architecture for an A-BPMS comprising five subsystems, as illustrated in \figurename~\ref{fig:abpms-architecture}: a data layer, a process intelligence layer, an action layer, an orchestration layer, and a conversational layer.


The \emph{data layer} integrates both structured and unstructured data about business operations, including event logs that record the execution of past and ongoing business processes, repositories containing process models, records of past (improvement) decisions, as well as other relevant data and documentation about the processes managed by the system.
This layer is fundamental for the A-BPMS to ground the management of the process on information about both the designed and the actual execution of the process.
By doing so, it allows the A-BPMS to ``perceive'' the current state and evolution of the business process, as well as its surrounding environment~\cite{dumas2023ai}.


On top of the data layer sits a \emph{process intelligence layer}, which provides the collection of techniques introduced in \sectionname~\ref{sec:from-process-mining-to-augmented}.
These techniques equip the A-BPMS with capabilities to ``explain'' the current and potential future states of the process, as well as to ``improve'' its performance and ``adapt'' to external changes through prescriptions or recommendations.


Next to the above layers, the \emph{action layer} provides capabilities for triggering actions that affect the execution of the process.
Such actions include, among others,
\textit{i)} creating or altering the state of a case in a business process, e.g., through a workflow management system;
\textit{ii)} interacting with external actors through collaboration tools, e.g., sending an e-mail to notify an applicant about a change in their application;
\textit{iii)} triggering a software bot to perform an automated activity; and
\textit{iv)} updating records in a CRM, ERP, or other Systems of Records.
This layer enables the A-BPMS to handle the execution of the process (``enact'').


The \emph{orchestration layer} rests on top of these layers, coordinating the sub-systems that manage the process and handling the decision-making to improve the process performance and adapt to external changes.
Composed of agentic and/or rule-based process orchestration systems, this layer allows the A-BPMS to ``reason'' based on the analytics provided by the process intelligence layer, and to effectively perform the actions enabled by previous layers.


Finally, the \emph{conversational layer} serves as an interface that exposes the capabilities of the system to various types of external agents and users.
The interaction between users and the A-BPMS is channeled through conversational agents powered by generative AI techniques -- e.g., Large Language Models.
Meanwhile, the interaction with external agents is implemented through \emph{Model Context Protocol} (MCP)~\cite{DBLP:journals/corr/abs-2503-23278} tools, which provide semantically rich descriptions for their consumption.


Within an A-BPM system, the data layer provides access to historical and current data for the techniques belonging to the process intelligence layer.
These techniques, which establish an interface layer to access the process data for all other components, generate insights that are consumed by the action layer, the orchestration layer, and the conversational layer.
The action layer utilizes these insights to make localized decisions -- e.g., when executing the process or planning future allocations --, closing the cycle by producing process data that is fed back to the data layer.
Meanwhile, the orchestration layer leverages the insights provided by process intelligence capabilities, as well as information from external agents and users, to coordinate the components of the action layer.
Finally, the conversational layer constitutes an interface that channels the exchange of information between the system and the external agents and users.

\section{Levels of Automation and Autonomy}

Traditionally, the execution of processes has been viewed as a linear spectrum ranging from manual to automated~\cite{zayas2021identifying,mishra2019people}.
The emergence of agentic AI presents a third and distinct mode: autonomous process execution~\cite{janiesch2025process,vu2025agentic}. 
In this section, we propose a conceptual model that extends the traditional human-to-automated spectrum with an autonomous dimension.
This model classifies BPM systems depending on both the actor that executes an activity and the actor that orchestrates the process, i.e., a human (manual), a rule-based system (automated), or an agentic AI (autonomous).

\subsection{The Process Execution Spectrum}

Activities within a business process (either atomic, semi-atomic, or long-running) can be executed by either a human, a rule-based bot, or an agentic AI.

The execution of an activity is considered manual when the main actor is a human, even when supported by autonomous or automated systems.
For instance, an employee registering an invoice in the system, even if assisted by a machine that automatically fills in the data, is considered a manual execution.

The automated execution of an activity is performed by rule-based bots that follow a fixed, pre-defined set of rules.
Their value lies in automating repetitive tasks where all possible scenarios can be anticipated and scripted.
For instance, a bot might be programmed to take an incoming invoice via email, extract the invoice data, and input the data into a system via an API~\cite{van2018robotic}.
In this context, no human interaction is needed.

The execution of an activity is considered autonomous when enabled by an agentic AI.
In this context, we define agentic AI as a software that has the capability to \emph{sense}, \emph{decide}, and \emph{act}.
``Sense'' refers to the ability to analyze its environment, process information from various sources -- such as software APIs, databases, or physical sensors --, and extract relevant insights.
``Decide'' refers to the capabilities to evaluate multiple options, reason through scenarios, and determine the next course of action.
``Act'' is the capability to execute the chosen strategy by, for instance, interfacing with external systems, coordinating with other agents, or directly instructing other systems to achieve its specified objectives without requiring human support or supervision.\footnote{
Agentic AI is distinctly different from generative AI.
While generative AI is capable of generating content once prompted, i.e., reactively, an agentic AI proactively performs actions and makes decisions in order to achieve a specific goal.
Agents are, therefore, autonomous.
}
For instance, an agentic system may be designed to assess the potential fraudulence of a loan application by autonomously executing a set of machine learning algorithms and combining their outputs with domain-specific knowledge.

Similar to the execution of activities, a business process can be orchestrated by a human, a rule-based system, or an agentic AI.
Here, the process orchestrator is the central entity responsible for guiding a business process from start to finish, defining the overall flow, managing dependencies, and ensuring the process achieves its objectives.

\begin{figure}[h]  
    \centering  
    \includegraphics[width=0.95\textwidth]{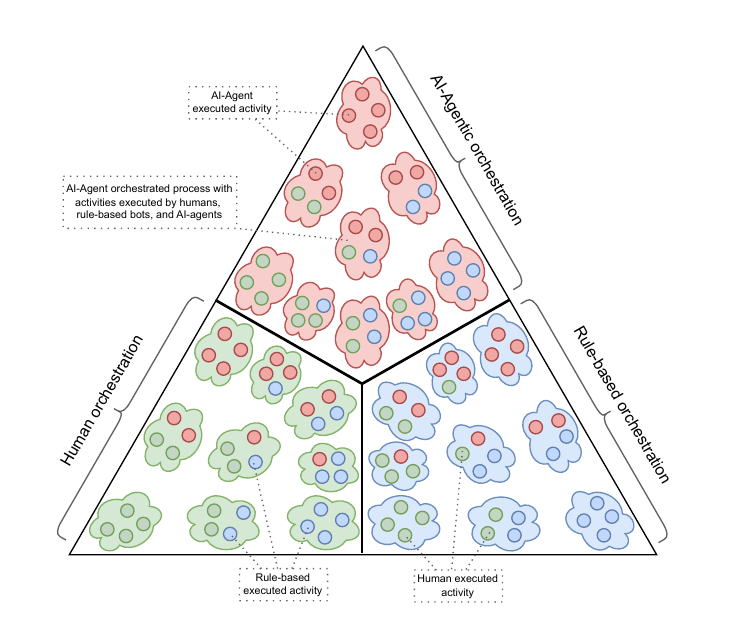}  
    \vspace{-25pt}
    \caption{Autonomy Spectrum of Business Process Execution.}  
    \label{fig:triangle}  
\end{figure}

In human orchestration, a human is the central conductor of highly variable and knowledge-intensive processes.
A rule-based orchestrator relies on predefined sets of rules or rigid workflow engines that execute deterministic processes automatically, a characteristic typical of traditional BPM suites, where the sequence of steps, decision points, and assignments are explicitly modeled upfront.
An agentic orchestrator represents the most advanced form, where an agent autonomously determines and manages the entire process flow to achieve a high-level goal.
We connect these foundational orchestrators into the (triangular) process execution spectrum depicted in \figurename~\ref{fig:triangle}, with three separated regions representing human, rule-based, and agentic orchestration of a business process.

At the bottom left region of the spectrum, we find processes orchestrated by human actors.
The bottom left vertex corresponds to traditional BPM, where human employees are responsible for executing all activities of the process from start to finish.
For instance, in criminal investigations~\cite{innes2002process}, this manual process is exemplified by the investigative framework, where detectives assess each situation, make decisions about resource allocation, conduct face-to-face interviews with victims and witnesses, and physically attend crime scenes to gather evidence, among other tasks.

At the bottom right region, we find processes orchestrated by rule-based systems.
The bottom right vertex represents processes that are both managed and executed by a predetermined set of rules and procedures.
This is the domain of traditional process automation.
The orchestration is procedural and fixed.
The system is designed to handle activities in a predictable and deterministic manner.
While the rules were designed by humans, their execution is automated. 
For instance, in manufacturing companies, such as Tesla, this is exemplified by robots that perform welding, painting, and assembly operations according to exact specifications.
Each robot executes its designated task in a predetermined sequence, following a program that dictates every movement, timing, and action~\cite{kaur2025robotics}.

Finally, at the top region of the triangle, we find agent-based orchestrated processes.
The top vertex represents fully autonomous process execution, where (AI-)agents orchestrate the process -- with the objective of achieving a defined high-level goal -- and execute all its activities.
For instance, in a procurement context, such AI agents autonomously conduct supplier negotiations.
They analyze market conditions in real-time, assess supplier behavior patterns, dynamically adjust negotiation strategies, and have the authority to finalize deals~\cite{herold2025brave}.

\subsection{Agentic Orchestration Patterns} 

Processes exist anywhere within this spectrum.
For instance, the detective (human orchestrator) may maintain control and decision-making over the process, but delegate some activities to rule-based systems (e.g., database searches for criminal records) or to agentic systems (e.g., analysis of digital evidence patterns).
As more automated or autonomous activity executions are included, the process position moves toward the bottom right (rule-based corner) or upward (agentic corner), but still within the human orchestration region of the triangle.

Within the boundaries of human, rule-based, or agentic orchestrator foundations, different types of agentic sub-orchestrators may emerge. 
Such patterns define how agents interact, delegate responsibilities, and manage workflows within a sub-process.
These patterns vary in structure, adaptability, and communication style, allowing system designers to tailor agent behavior to specific use cases.

The most common and standard type would be \emph{Sequential Orchestration}, where agents operate in a fixed pipeline.
Sequential Orchestration involves arranging agents in a fixed linear pipeline where each agent performs a specific task and passes its output to the next agent in the sequence.
For instance, in contract creation, the process might begin with a template selection agent, followed by a clause customization agent, then a compliance checker, and finally a risk assessment agent.
Each agent adds value incrementally, ensuring a thorough and structured output.

A variation and slightly more complex type is the \emph{Parallel Orchestration}, which enables multiple agents to operate concurrently, either on the same task or on different aspects of a task simultaneously.
For instance, during product evaluation, technical, legal, and business agents can independently assess the product from their respective domains, providing a comprehensive and multi-faceted analysis that would be slower or less rich if done sequentially.

The next level of complexity requires some form of delegation where one agent acts as a supervisor or a manager.
Here, the \emph{Routing Pattern} features a central agent that routes tasks to specialized agents.
This pattern would be more common in customer support systems, where a supervisor agent might route billing inquiries to a finance agent, technical issues to a support agent, and legal questions to a compliance agent.
It ensures efficient handling of diverse queries while maintaining centralized oversight.

A variation is the \emph{Managerial Orchestration}, where a master agent works with task delegation.
Managerial Orchestration features a master agent that decomposes a complex task into subtasks and delegates them to specialized sub-agents.
The master agent ensures that subtasks are executed in the correct order and that dependencies are respected.
For instance, in a travel planning system, the master agent might assign hotel booking to one agent, flight search to another, and itinerary creation to a third, ensuring that all components align with the user’s preferences and constraints.

When agents themselves decide when and to whom to transfer the work, we have \emph{Adaptive Orchestration}, which allows agents to transfer control based on context.
Adaptive Orchestration focuses on dynamic delegation, where agents transfer control to others based on the evolving context of a task.
This pattern is useful in workflows that require adaptability and context-aware decision-making.
An example could be a summarization agent that, after generating a summary, hands off the output to a visualization agent to create a graphical representation.

The next level of complexity is when agents can collaborate without direct supervision.
For instance, \emph{Mesh Orchestration} supports decentralized peer-to-peer coordination.
In this model, agents communicate directly with one another, passing tasks off as needed without the need for a central controller.
For instance, agents collaboratively resolving a support ticket might pass the issue from one to another based on expertise, availability, or context, adapting dynamically to the situation without relying on a central orchestrator.

Finally, the most advanced type is the \emph{Self-orchestration}, where agents self-select tasks based on relevance.
Self-orchestration introduces a mechanism where tasks emit signals that attract relevant agents.
Agents evaluate these signals and self-select tasks based on their capabilities, availability, or interest.
For instance, in a report analysis system, agents might self-assign to different sections of a document based on their domain expertise, ensuring that each part is handled by the most qualified agent.

Each pattern offers distinct advantages depending on the complexity, adaptability, and control requirements of the task at hand.
The choice of pattern and implementation of Agentic AI should be determined from the nature of the business process that is to be supported or replaced.
However, these patterns form a versatile toolkit for designing agentic AI systems that can support virtually all kinds of business processes.

\subsection{Agentic Execution Patterns}

In the execution of sub-processes and activities, different patterns emerge on how humans, rule-based systems, and agentic AI interact at various levels of granularity, including triage, human-assisted, agent-assisted, and verification patterns.

The \emph{Triage Pattern} involves intelligent decision-making that dynamically routes cases or activities to the most appropriate executor -- human, rule-based, or agent-based.
This pattern optimizes resource allocation by ensuring that each activity is handled by the most suitable actor.
For instance, in a customer service process, incoming inquiries are automatically assessed and routed.
In this context, simple inquiries may be handled autonomously by AI; moderately complex issues that can be solved by executing predefined steps -- e.g., billing disputes -- may be processed by rule-based systems; and sensitive complaints may be escalated to human actors.

\emph{Human-Assisted Agent} represents a pattern where an agent performs the primary activity, autonomously making decisions, while a human provides oversight, verification, or final approval to ensure quality, compliance, and accountability.
This pattern is relevant, for instance, in high-stakes environments where human accountability remains non-negotiable.
For instance, an AI agent might analyze market data, process complex algorithms, and generate investment recommendations for private banking customers.
However, the client executive must review, validate, and approve these recommendations before they are executed.

\emph{Agent-Assisted Human}, on the other hand, involves humans performing the primary activity or decision-making, while AI agents provide verification, quality assurance, additional insights, or real-time support to enhance human performance and reduce errors.
For instance, in customer support operations, human representatives may handle complex customer calls that require empathy, problem-solving, and relationship management, while AI agents simultaneously analyze the conversation to assess factors such as tone, politeness, compliance with company policies, and customer satisfaction indicators.

The \emph{Verification Pattern} establishes quality control mechanisms in which one type of performer systematically reviews, checks, or validates the output of another to ensure accuracy, compliance, and adherence to standards.
An example of this pattern occurs in content creation workflows, where a human writer may produce initial content leveraging their creativity and domain expertise, while an AI agent conducts a comprehensive review for factual accuracy, tone consistency, and alignment with brand guidelines.

\section{Implications for Research}

The attention and traction of LLM models and Agentic AI might require revisiting traditional BPM, similar to how the maturation of client-server architectures impacted BPM. Earlier automation waves considered a spectrum of manual and rule-based execution. However, agentic performers introduce autonomous software that can sense, decide, and act beyond pre-programmed rules. This can open up the field for new research opportunities.

One area of research is in the process modelling stage. Today’s activity-centric notations might prove inadequate once “agent” becomes a first-class performer. A purchase-to-pay model written in BPMN, for instance, lacks constructs for expressing an agent’s planning capability or the constitutional frame that constrains it. Future formalisms should support, for instance, objective-based blocks, guard-rail annotations, and verification patterns. 

The introduction of agentic AI systems, with their autonomous decision-making capabilities and complex human-AI interactions, necessitates the inclusion of verification-centric process design where quality control, accuracy validation, and compliance checking become part of the design phase. Research is needed to develop frameworks that identify where verification patterns should be embedded within processes, how to design multi-layered validation mechanisms that can operate across human, rule-based, and agentic components, and methods for ensuring that autonomous agents maintain accountability.

Organisations will gradually redesign existing processes to include LLM and agentic AI components. However, history has shown that technology substitution creates modest gains. Client-server platforms and the early Internet made broad automation possible, yet real value emerged only when firms redesigned work to exploit once-only data capture, automated rule checking, and web self-service. LLM and agentic AI follow the same pattern. By the same logic, it might be valid to consider what business process improvement opportunities LLM and agentic AI can enable. Hence, the existing catalogue of redesign heuristics (task elimination, parallelism, integral technology) might need an agentic extension.

\bibliographystyle{splncs}
\bibliography{references}

\end{document}